\documentclass{article}

    \PassOptionsToPackage{numbers, compress}{natbib}

\usepackage[preprint]{neurips_2023}

\usepackage[utf8]{inputenc} 
\usepackage[T1]{fontenc}    
\usepackage{url}            
\usepackage{booktabs}       
\usepackage{amsfonts}       
\usepackage{nicefrac}       
\usepackage{microtype}      
\usepackage{xcolor}         
\usepackage{graphicx}       
\usepackage{tabularx}
\usepackage{wrapfig}
\usepackage{pythonhighlight} 
\usepackage{notoccite}
\usepackage[bookmarks=false]{hyperref}

\title{Breaking the Token Barrier: Chunking and Convolution for Efficient Long Text Classification with BERT}

  \author{Aman Jaiswal \\
  Dalhousie University, Halifax \\
  \texttt{aman.jaiswal@dal.ca} \\\And
    Evangelos Milios \\
  Dalhousie University, Halifax \\ 
  \texttt{evangelos.milios@gmail.com}}

\begin{document}

\maketitle

\begin{abstract}
Transformer-based models, specifically BERT, have propelled research in various NLP tasks. However, these models are limited to a maximum token limit of 512 tokens. Consequently, this makes it non-trivial to apply it in a practical setting with long input. Various complex methods have claimed to overcome this limit, but recent research questions the efficacy of these models across different classification tasks. These complex architectures evaluated on carefully curated long datasets perform at par or worse than simple baselines. 
In this work, we propose a relatively simple extension to vanilla BERT architecture called ChunkBERT that allows finetuning of any pretrained models to perform inference on arbitrarily long text. The proposed method is based on chunking token representations and CNN layers, making it compatible with any pre-trained BERT. We evaluate chunkBERT exclusively on a benchmark for comparing long-text classification models across a variety of tasks (including binary classification, multi-class classification, and multi-label classification). A BERT model finetuned using the ChunkBERT method performs consistently across long samples in the benchmark while utilizing only a fraction (6.25\%) of the original memory footprint. These findings suggest that efficient finetuning and inference can be achieved through simple modifications to pre-trained BERT models.
\end{abstract}

\section{Introduction}

Transformers \cite{DBLP:journals/corr/VaswaniSPUJGKP17} and their derivatives, e.g, BERT \cite{DBLP:conf/naacl/DevlinCLT19}, RoBERTa \cite {DBLP:journals/corr/abs-1907-11692} have achieved significant improvements in various NLP tasks including text classification. These improvements have been largely credited to the self-attention module of the architecture. Self-attention, also known as intra-attention, is a mechanism used in the BERT model (and other Transformer models) to allow the model to attend to different parts of the input text at the same time, rather than processing the text sequentially. This contrasts previous models such as recurrent neural networks (RNNs \cite{GilesKW94}), which process input sequentially, one word at a time. This allows the model to capture long-range dependencies in the input text more effectively.
 
The self-attention mechanism in BERT has a fixed-size window, which means it can only attend to a limited number of words at any given time. The exact size of this window varies depending on the specific architecture of the BERT model. Still, in many cases, it is limited to 512 tokens (i.e., words or sub-word units) in the input. This means that BERT is not able to process input sentences that are longer than 512 tokens. There are several reasons why the self-attention mechanism in BERT is limited to 512 tokens in the input. One reason is computational efficiency. Self-attention mechanism grows quadratically with the input length, which can be computationally intensive, and limiting the number of tokens in the input allows BERT to process the input more efficiently. By limiting the input to 512 tokens, BERT is able to focus on the most relevant words in the input and generate high-quality contextualized representations of those words. The limitation of BERT to 512 tokens in the input is a trade-off that allows the model to achieve good performance on a wide range of natural language processing tasks while also being computationally efficient.
In practice, it is non-trivial to apply BERT to real datasets which are much longer than the allowed limit of various off-the-shelf pre-trained models \cite{DBLP:journals/corr/abs-1910-03771}. This presents a challenge in terms of real-life application and feasibility. Different approaches have claimed to overcome this input limit but recent research from \cite{DBLP:journals/corr/abs-2203-11258} shows that when compared against the same benchmark these models have performed marginally better or even fail to outperform simple baselines (that truncates longer text) on different datasets. We propose ChunkBERT, a finetuning extension to vanilla BERT that can be used to finetune any pre-trained BERT on longer text. The chunkBERT finetuning strategy has the following advantages:
\begin{itemize}
    \item It  can adapt any off-the-shelf pre-trained BERT beyond the limit imposed by architecture.
    \item It allows fine-tuning on a standard GPU and the ability to scale down memory requirements if needed. In our setting, we reduce memory by a factor of 16, i.e., 6.25\% of the original memory.
    \item It scales linearly and can infer on inputs of arbitrary length, without the need for custom CUDA kernels.
    \item It is more systematic than random text truncation.
\end{itemize}

\section{Related Work}
There are a few different ways to extend BERT to handle input sequences that are longer than 512 tokens. One way is to use a variant of BERT that has a larger self-attention window (e.g, XLNet \cite{DBLP:journals/corr/abs-1906-08237}). This includes pretraining a model from scratch on in-domain data and finetuning on the downstream task, requiring a lot more computing than a simple finetuning task. Another way to extend BERT to handle longer input sequences is to use a "hierarchical" approach as \cite{DBLP:conf/cikm/Yang00BN20}, where the input sequence is divided into multiple shorter sequences and each of these sequences is processed separately by BERT.
 Another approach is to use BERT as part of a larger model that is designed to handle sequential inputs \cite{DBLP:conf/acl-mrqa/ChanF19}. 
Generally, there are four main approaches that can be broadly identified for handling longer sequences in the transformer architecture. The straightforward approach is to simply truncate the longer inputs (until 512 tokens) to fit within the allowed input limit of the architecture, this is the simplest and most popular approach. The disadvantage of simple truncation is that important information for the task may be truncated. A better approach can be to find the important parts of the longer input sequence and truncate the least important parts, this is called key-sentence selection. \cite{DBLP:conf/nips/DingZY020} explores this method by training two models in tandem, where the first model recognizes the important parts of the longer input and the second model inferences over the selected shorter input. Other approaches (e.g, Bigbird \cite{DBLP:conf/nips/ZaheerGDAAOPRWY20}, Longformer \cite{DBLP:journals/corr/abs-2004-05150}) focus on  efficient versions of the self-attention module which employ sparse self-attention and save memory by not attending to every token in the input. The hierarchical processing of longer documents \cite{DBLP:conf/asru/PappagariZVCD19} explores dividing longer inputs into shorter chunks and combining these chunks representation using an additional transformer attention layer.

More recent work \cite{DBLP:journals/corr/abs-2203-11258} highlights the lack of long-text classification benchmark and compares the above approaches on a unified benchmark across different classification tasks and datasets. The long-text benchmark \cite{DBLP:journals/corr/abs-2203-11258} suggests strong baselines that often outperform the state-of-the-art in terms of classification performance. Their baselines use the key-sentence selection strategy  where a partition of the long input is selected randomly or by using a text ranking algorithm.
\begin{figure} [t]
    \centering
    \includegraphics[width=\textwidth]{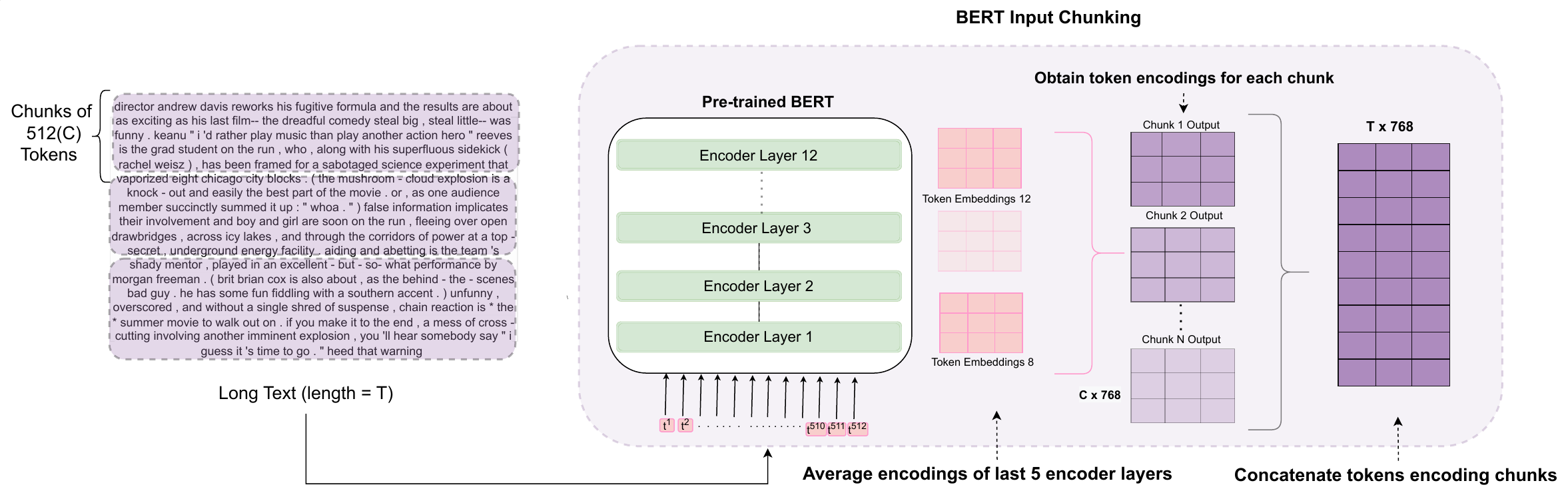}
    \caption{Overview of BERT Input chunking. First, the complete input $(T)$ is tokenized using the BERT tokenizer and split into chunks of size $(C)$ e.g, 128 tokens each. The input chunks are independently processed by BERT and produce token embeddings for each chunk of size $(C \times 768)$. Here, we average the token embeddings of the last five layers of the BERT model to obtain token representations. Once every chunk is processed, token representations are concatenated over the token to produce the complete input matrix of size $(T \times 768)$. The input matrix is used by TextCNN classification module.}
    \label{fig:bert-chunking}
\end{figure}

The success of simple baselines against complex models motivated our shift in perspective to investigate finetuning strategies rather than compute-efficient attention mechanisms, this work is parallel to efficient attention mechanisms and complements any global attention approximations (e.g, Nyströmformer \cite{DBLP:journals/corr/abs-2102-03902}). Our work focuses on taking advantage of the transfer learning paradigm (pretaining-finetuning) and reusing pretrained models for finetuning on longer text.

\section{ChunkBERT}
We propose a simple chunking methodology to consume long inputs, where each chunk is processed independently and their representations are concatenated to be processed by a TextCNN classification module. The complete architecture is optimized end-to-end and learn to use the chunk representation to perform well in the downstream task. The finetuning can be divided into 4 steps: 
\begin{itemize}
    \item Dynamic padding of batches to support splitting of batched inputs into equal-sized chunks.
    \item Splitting the long input into shorter input chunks and independent processing of each chunk.
    \item Concatenation of chunk representation to obtain the complete input embedding.
    \item Classification using the TextCNN module.
\end{itemize}

We achieve this using the token representation induced by the BERT model and chunking them until the complete input is observed. Specifically,  The long input of size $(T)$ is split into digestible chunks of size $(C)$ giving $(\frac{T}{C})$ or $(n)$ chunks of input tokens. BERT processes each input chunk independently to induce token representations of size $(C \times 768)$. The token representations from each split $(n \times (C \times 768))$ are concatenated to obtain the final token-level representations of the complete input $(T \times 768)$. The BERT input chunking methodology is illustrated in Figure \ref{fig:bert-chunking}.

Finally, we process the concatenated embeddings obtained by BERT input chunking using a TextCNN module as described in \cite{DBLP:conf/emnlp/Kim14}. The whole pipeline of BERT chunking and CNN layer is trained end-to-end using cross-entropy loss. This allows us to process arbitrary long sequences using the constant parameters of the CNN layers. 

\subsection{Chunk Attention}
 \begin{wrapfigure}{l}{0.5\textwidth}
    \begin{center}
    \vspace{-0.30in}
	\includegraphics[width=0.45\textwidth]{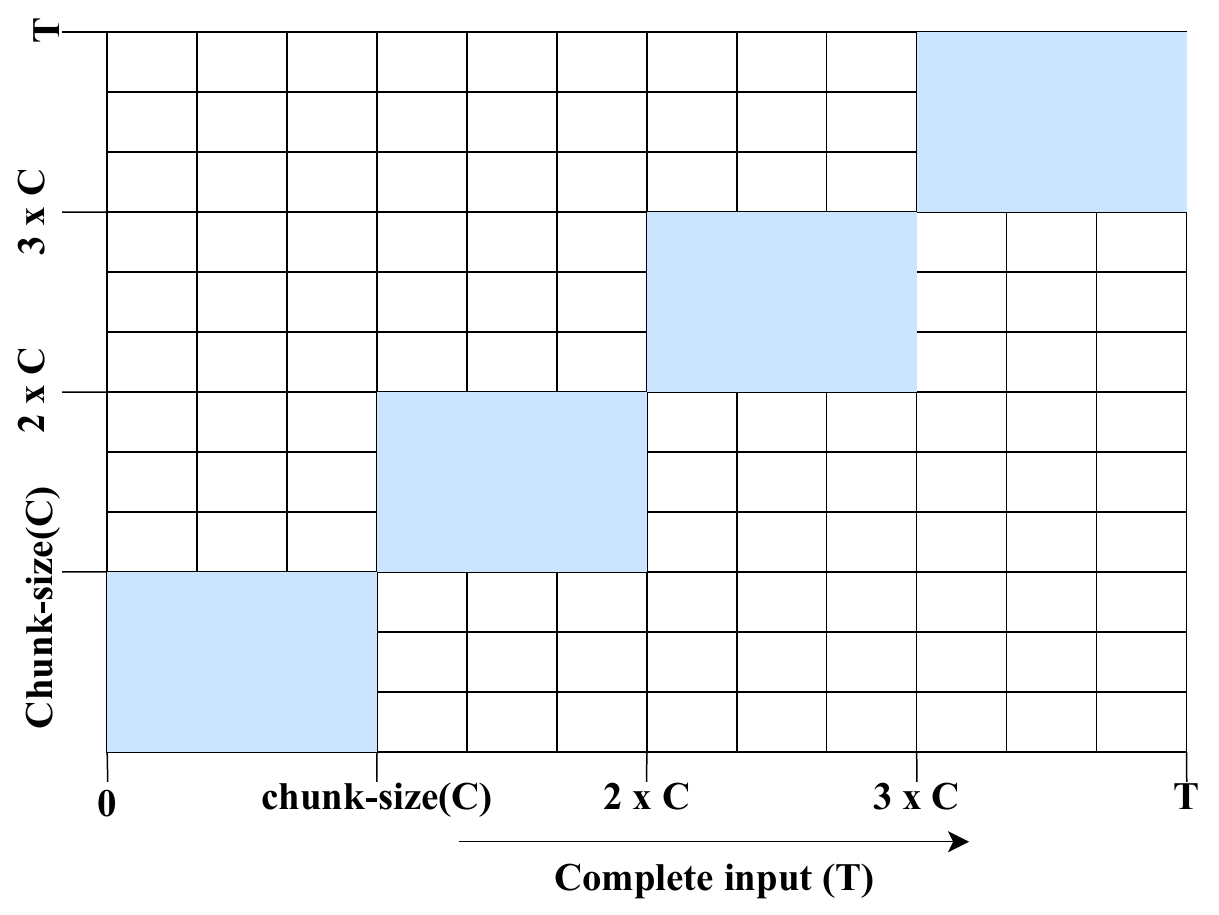}   
    \end{center}
    \vspace{-0.20in}
    \caption{Chunk Attention Mask: illustration of chunk attention for the complete input $T$. The space complexity becomes $O(nC^2)$, where $n$ is number of chunks and $C$ is the chunk-size.}
    \label{fig:chunk-attention}
\end{wrapfigure}

The chunking mechanism is more efficient than global attention as it only computes local attention weights and routes inter-token information within each chunk independently. This makes the space complexity quadratic to chunk-size$(C)$ rather than input length $(T)$ and linear to the number of chunks $(n)$. This means that we can potentially increase the number of chunks for a linear growth in space requirement while consuming longer inputs. The chunk attention mask is illustrated for a complete input in Figure \ref{fig:chunk-attention}. It highlights the need for routing inter-chunk information. In classification tasks, it can be important for information in the first chunk to be routed to the last chunk in the input. We model an aggregation layer to combine this inter-chunk information before classification using a TextCNN module.

\subsection{ Vectorized Chunking}
The process of inducing input chunk representations may be done sequentially, where one forward pass is needed for each chunk. The sequential method is suitable when memory is limited, allowing finetuning and inference in low-resource settings. Since each chunk only looks at the local context of the chunk window, this process can be vectorized by shifting the chunks into the batch dimension and obtaining the complete token representation in a one-forward pass. The chunk outputs can be shifted back into the token dimension before being processed by the TextCNN module. This process is facilitated by dynamic padding of the inputs to the next multiple of chunk size ($C$). Vectorized chunking allows for even faster fine-tuning and inference when resources are available. The pseudo-code for vectorized chunking is shown in Appendix \ref{appendix:b}

\subsection{TextCNN Aggregation}

\begin{figure}[ht]
    \centering
    \includegraphics[width=0.5\textwidth]{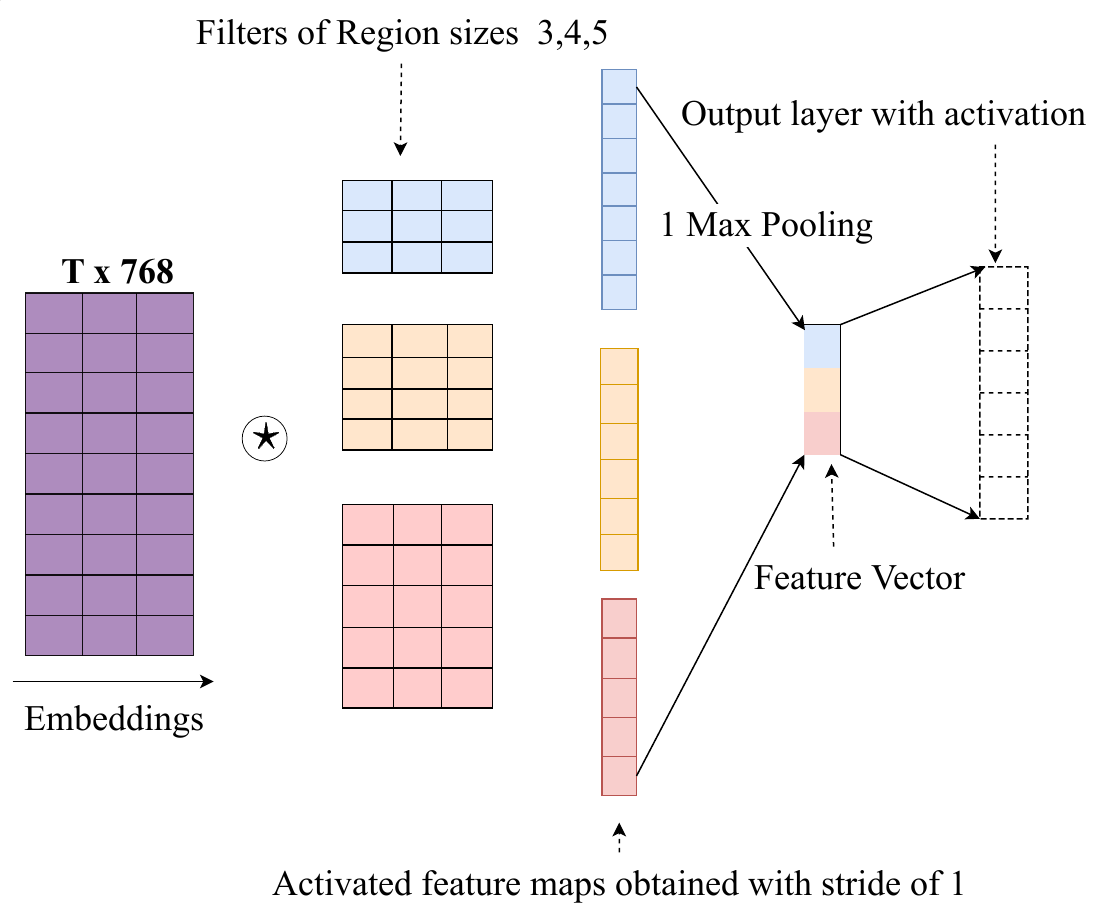}
    \caption{TextCNN architecture from \cite{DBLP:conf/emnlp/Kim14}, We use 100 filters of window sizes 3, 4, and 5 each which slide over the input $(T \times 768)$ with stride 1. This is followed by a max-pooling operation to obtain the feature vector that is projected to the classification layer. \textit{Note:} The illustration only shows one filter for each window. The final feature vector is of size 300, obtained by concatenating 100 filters for each window.}
    \label{fig:chunk-bert-cnn}
\end{figure}

The concatenated token representations $(T \times 768)$ can be used as input for another module that can combine information from each chunk to give an aggregated comprehension of the input. We use a TextCNN module for this aggregation task, as it can process  concatenated token representation of any length ($T$) with the fixed amount of parameters it learned  during the finetuning stage. TextCNN is a type of convolutional neural network (CNN) designed for natural language processing tasks, such as text classification and sentiment analysis. \cite{DBLP:conf/emnlp/Kim14} proposed the Convolutional neural networks for text processing where they used pretrained word vectors as input. The TextCNN model uses multiple filters of different sizes to process the input text and extract features, which are then fed into a fully connected layer to make predictions. The filters are applied to the input text in a sliding window fashion, allowing the model to capture local dependencies within the text. The TextCNN model used in ChunkBERT is illustrated in Figure \ref{fig:chunk-bert-cnn}.

\section{Evaluation}

\subsection{Benchmark}
We used the long text benchmark proposed by \cite{DBLP:journals/corr/abs-2203-11258}, which provides a comprehensive comparison of existing models for long document classification across a range of unified datasets and baselines. The benchmark covers different models representing various strategies for modeling long-text classification in BERT. It includes various datasets, such as binary classification, multi-class classification, and multi-label classification. Table \ref{dataset-stats} presents the statistics of the datasets with their task types.

\subsection{Datasets}

We briefly describe the datasets used in the benchmark, including the artificially modified dataset to increase their length in Table \ref{table:dataset-desr}. Since not all samples in a dataset may be above the context length of 512 tokens, it is essential to highlight the relative percentage and the absolute number of samples longer than the context length, as shown in Table \ref{dataset-stats}.

\begin{table}[!hbt]
\centering
\caption{Dataset descriptions in the long text benchmark \cite{DBLP:journals/corr/abs-2203-11258}}
\begin{tabularx}{\textwidth}{@{}lX@{}}
\toprule
Dataset & Description\\ \midrule

Hyperpartisan & A binary classification dataset introduced in SemEval-2019 \cite{hyperpartisan}. Each news article can either be 'hyperpartisan' or 'not hyperpartisan.' The term hyperpartisan refers to an extreme and uncompromising bias toward a particular political party or ideology. \\ \midrule
20NewsGroups &  A popular multi-class dataset \cite{20news}, where each news item can belong to one of the 20 news categories. \\ \midrule
Eurlex-57K & A multi-label classification dataset with 4.3K EUROVAC labels based on the European Union's legislation documents, originally proposed for zero-shot or few-shot learning \cite{eurlex}. It consists of three sections, with the first two sections' headers and recitals being the most information-dense. This dataset has the maximum number of long samples in the benchmark, with 3078 samples.\\ \midrule
Inverted-Eurlex & A simulated version of the EURLEX-57K dataset \cite{eurlex}, where the salient sections (headers, recitals) are shifted toward the end of the document after the main body. This approach should enforce machine learning models to observe and retain the information until the end of the document to make correct predictions.\\ \midrule
CMU Book Summary & A multi-label dataset \cite{cmubooksummary} extracted from Wikipedia, which contains book summaries and their metadata, such as author and genres from Freebase. The genres are used as labels.\\ \midrule
Paired Book summary & A simulated dataset made from the CMU book summary \cite{cmubooksummary}, where two summaries are joined to create a longer text with two independent information blocks. This requires machine learning models to predict the correct genres for both summaries. This dataset has the longest text, with a mean of 1147 tokens. \\ 
\bottomrule
\end{tabularx}
\vspace{1pt}

\label{table:dataset-desr}
\end{table}

\begin{table}[!htb]
\centering
\caption{Dataset statistics and task types. B, MC, and ML refer to binary, multi-class, and multi-label respectively. \#BERT tokens are the average token counts after tokenization using a BERT-base tokenizer. \# Long refers to the number of long samples in the test-set. Note: The dataset statistics are reproduced from \cite{DBLP:journals/corr/abs-2203-11258}}
\label{dataset-stats}
\begin{tabularx}{\textwidth}{@{}*{8}l@{}}
\toprule
Dataset       & Task~       & \# Train & \# Dev & \# Test & \# Labels & \# BERT Tokens & \# Long (\%)\\
\midrule
Hyperpartisan & B     & 516      & 64     & 64      & 2         & 744 ± 677     & 34\space\space\space\space\space(53\%) \\
20NewsGroup   & MC & 10,182   & 1,132  & 7,532   & 20        & 368 ± 783     & 1107 (15\%)\\ \cmidrule(r){1-1}
EURLEX-57K    & ML & 45,000   & 6,000  & 6,000   & 4271      & 707 ± 538     & 3078 (51\%)\\
~-Inverted    & ML  & 45,000   & 6,000  & 6,000   & 4271      & 707 ± 538     & 3078 (51\%)\\ \cmidrule(r){1-1}
Book Summary  & ML & 10,230   & 1279   & 1279    & 227       & 574 ± 659     & 495\space\space\space(39\%)\\
- Paired      & ML & 5115     & 639    & 639     & 227       & 1147 ± 933    & 482\space\space\space(76\%) \\
\midrule
\end{tabularx}
\end{table}

\subsubsection{Baseline models}
\begin{figure}[!ht]
    \centering
    \includegraphics[width=\textwidth]{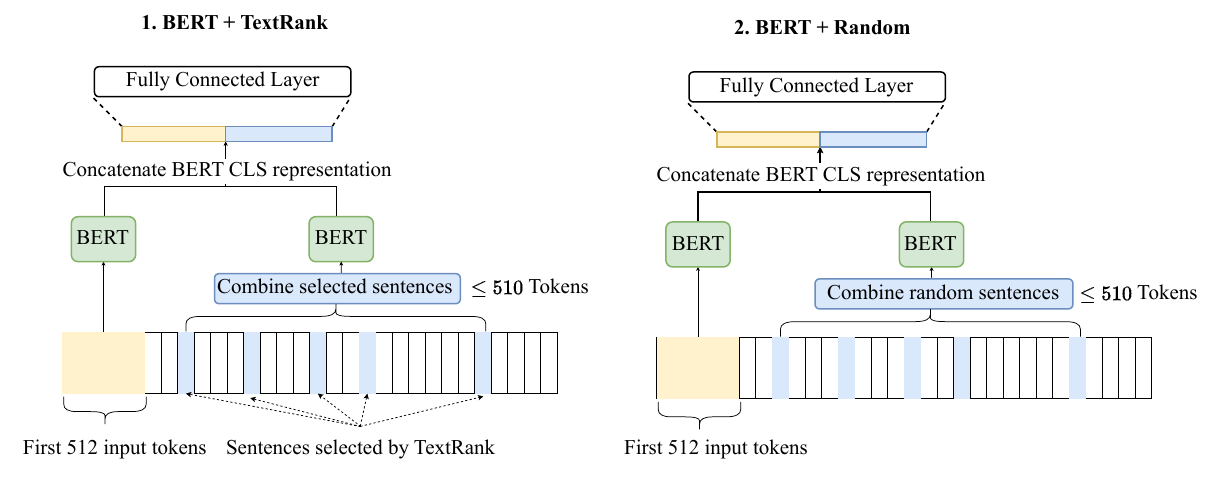}
    \caption{Baselines: 1) BERT+TextRank concatenates the CLS token embedding of the first 512 tokens with another 512 tokens obtained by a text ranking algorithm \cite{textrankalgo}. This concatenated $2 \times 768$ embedding is used for the output layer. 2) BERT+Random augments the first 512 tokens with random 512 tokens. \textit{Note}: The baselines are originally described in \cite{DBLP:journals/corr/abs-2203-11258}.  }
    \label{fig:baselines}
\end{figure}

The benchmark study \cite{DBLP:journals/corr/abs-2203-11258}, proposed simple yet effective baseline models that outperformed complex architectures. We will briefly describe these baseline models, using our own illustrations in Figure \ref{fig:baselines}. The first baseline model is called BERT+Textrank, which uses an off-the-shelf implementation \citep{pytextrank} of an unsupervised sentence ranking algorithm \citep{textrankalgo} to extract key sentences from the input text. The [CLS] representation of the first 512 tokens and the key sentences (up to 510 tokens) are concatenated and projected to the output layer. This requires two forward passes of the BERT model. The second baseline model is called BERT+Random, which is similar to the first model but instead of using key sentences, it selects random 512 tokens from the input text.

The benchmark study also includes three additional models that cover different strategies for modeling long texts. The first model is the Longformer \citep{DBLP:journals/corr/abs-2004-05150} model, which uses efficient sparse attention to model long-range dependencies. The second model is ToBERT \citep{DBLP:conf/asru/PappagariZVCD19}, which stands for transformer over BERT. It splits the input into overlapping segments and applies a 2-layer transformer over their representation. The third model is CogLTX \citep{DBLP:conf/nips/DingZY020}, which uses key sentence selection to train two models in tandem, one for selecting relevant input sentences and the other for classification using the selected input.
\subsection{Experimental Setup}

During training, we may limit the number of maximum input tokens to allow for efficient training. Even though chunkBERT allows for arbitrary input length for both training and testing, we set this to 4096 tokens, which is more than sufficient for the input in the considered datasets. There are two reasons for this: first, to limit the memory requirement during training, and second, to match the input limit of the LongFormer model. The chunk size ($C$) is set to 128 tokens to reduce the memory consumption to 6.25\% of the original BERT with 512 context length. We found Chunksize of 128 tokens provides a good tradeoff between training efficiency and performance. The TextCNN module and the underlying BERT model share the same ADAM optimizer and learning rate of 3e-05 with a batch size of 8. The model is trained for 20 epochs using early stopping on the development set and evaluated on the test set as followed by \cite{DBLP:journals/corr/abs-2203-11258}. We report the average of evaluation metrics over 5 runs. We utilize and modify the original codebase\footnote{\url{https://github.com/amazon-science/efficient-longdoc-classification}} for our experiments. 

\subsection{Results}
The evaluation metrics for all models, except for ChunkBERT, are reported directly from \cite{DBLP:journals/corr/abs-2203-11258}. Table \ref{onlylong}. displays the experimental results on the test set containing only long samples (i.e., samples with more than 512 tokens). The table shows that ChunkBERT and Longformer exhibit similar performances on the Hyperpartisan dataset. The baseline models show the best performance in three out of the six datasets. LongFormer's performance varies across datasets, while ChunkBERT consistently performs well across all datasets, albeit slightly behind the best-performing models. The ChunkBERT model performs comparatively with only 128 chunk sizes i.e., with a much smaller footprint. On average, across all datasets, ChunkBERT emerges as the best-performing model. It performs 5.7\% better than Longformer on average, and about 18\% better in complicated datasets like EURLEX and Inverted-EURLEX, which have 4271 classes with many classes having zero samples. Similar trends are observed in the experimental results on the complete test set, as described in Appendix \ref{appendix:a}. The original baseline models continue to display a strong competitive performance highlighting the need to further study the benchmark datasets.
\begin{table*}[!thb]
    \centering
        \caption{Evaluation metrics on long documents (any document above 512 tokens) in the test set for all datasets. The average accuracy(\%) over five runs is reported for Hyperpartisan and 20NewsGroup while the average micro-F1 is reported for other datasets. The highest value per column is in bold. ToBERT on 20NewsGroups was omitted in the original table, which requires further preprocessing. The \textit{Average} column describes the average across all the datasets. \textsuperscript{*} ToBERT average performance does not include 20NewsGroups.\\}
    \label{onlylong}
    \resizebox{\textwidth}{!}{%
    \begin{tabularx}{\textwidth}{@{}lXXXXXXX@{}}
    \toprule
        Model  & Hyper-partisan & 20News-Group & EURLEX & Inverted-EURLEX & Book-Summary & Paired-Summary & Average \\
        \midrule 
        BERT & 88.00 & 86.09 & 66.76 & 62.88 & 60.56 & 52.23 & 69.42 \\ 
        BERT+TextRank & 85.63 & 85.55 & 66.56 & 64.22 & 61.76 & 56.24 & 69.99 \\ 
        BERT+Random & 83.5 & 86.18 & \textbf{67.03} & \textbf{64.31} & \textbf{62.34} & 56.77 & 70.02 \\ \midrule
        LongFormer & \textbf{93.17} & 85.5 & 44.66 & 47.00 & 59.66 & \textbf{58.85} & 64.81 \\ 
        ToBERT & 86.5 & NA & 61.85 & 59.50 & 61.38 & 58.17 & 65.48\textsuperscript{*} \\ 
        CogLTX & 91.91 & 86.07 & 61.95 & 63.00 & 60.71 & 55.74 & 69.9 \\ 
        \midrule
        \textbf{ChunkBERT} & 93.00 & \textbf{86.62} & 64.94 & 62.94 & 57.80 & 57.73 & \textbf{70.51} \\ \bottomrule
    \end{tabularx}%
    }
\end{table*}

\begin{figure}
    \centering
    \includegraphics[width=0.6\textwidth]{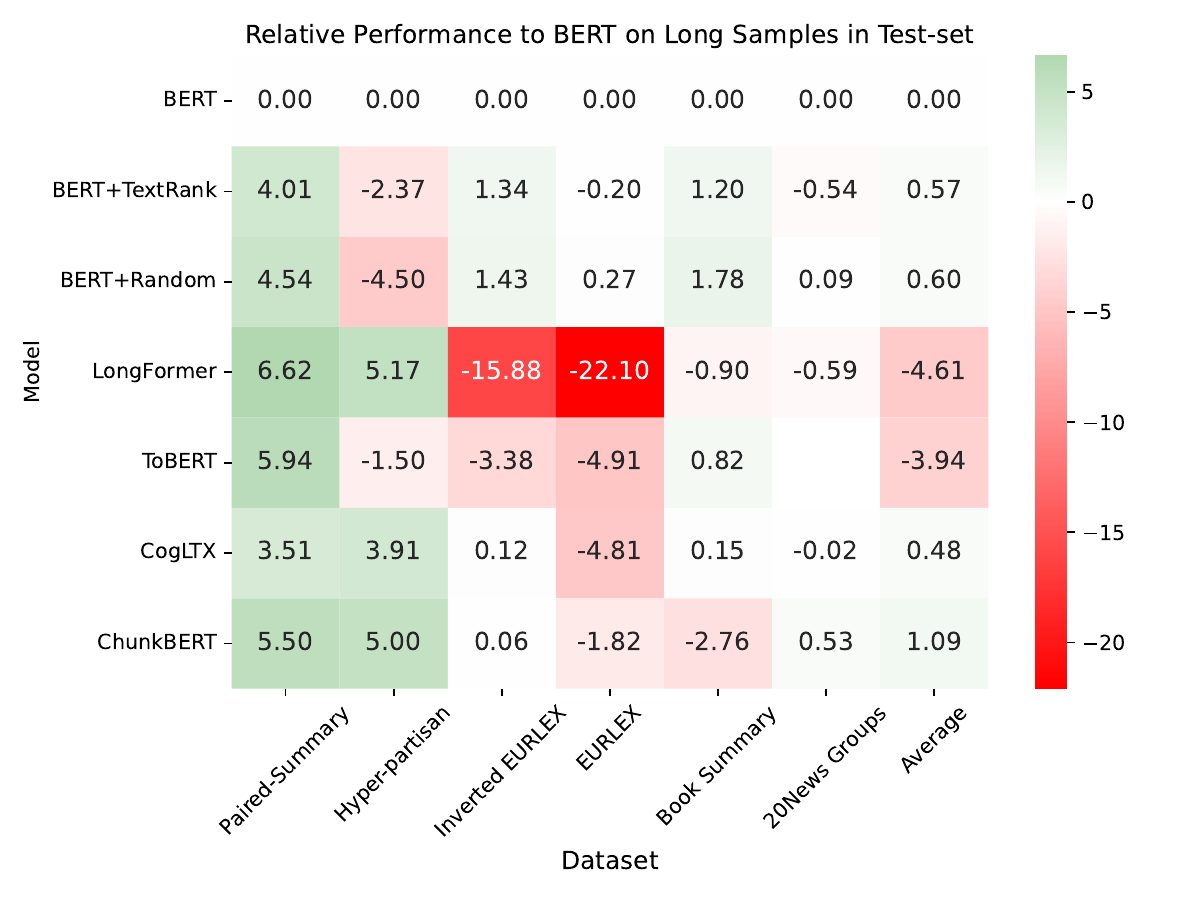}
    \caption{A heatmap displaying the performance metric difference ($\Delta$) between the BERT model and other models over the long documents ($> 512$ tokens) in the test set.. The x-axis is sorted by the percentage (\%) of long documents in the dataset. It provides a visual comparison of performance gain or loss relative to BERT.}
    \label{fig:rellong}
\end{figure}

\section{Discussion}
In this section, we discuss the results in terms of the location of discriminatory information in the dataset and the impact of different long-text modeling strategies on classification performance compared to simple BERT with truncation. This can be achieved by inspecting the performance difference ($\Delta$) between BERT and other models, as shown in Figure \ref{fig:rellong}. A positive delta suggests that there is significant discriminative information located only after the first 512 tokens, and the model is capable of utilizing this information effectively. On the other hand, a negative delta indicates that the model was unable to leverage the information even within the first 512 tokens, possibly due to the chosen modeling strategy.

In simple datasets, where salient information for correct classification is located within the first 512 tokens. For example, in the 20newsgroup dataset, all models perform similarly as the news category can be resolved from the first few sentences. Conversely, in the paired book summary dataset, all the models perform better than simple BERT as the models must consider the complete input to correctly classify the genres for both summaries in the document. Introducing unrelated sentences without full context may introduce noise and degrade performance, as observed in the hyperpartisan dataset, where BERT outperforms both the BERT+TextRank and BERT+Random alternatives. 

Complex datasets may require the complete interaction of tokens to accurately classify the documents. This is evident with the EURLEX dataset and its inverted variant, where the sparse attention of Longformer has a notable negative impact on performance compared to BERT. However, other modeling strategies are less affected, with ChunkBERT exhibiting the least degradation in performance. The Eurlex documents have key information concentrated at the beginning, which BERT leverages more effectively than complex models using full self-attention. Therefore, incorporating additional information to BERT using methods like TextRank or Random does not necessarily lead to significant performance improvements. However, in the inverted-Eurlex dataset, where the key information is shifted towards the end of the document, these variants do show improvements over BERT.

In general, the ChunkBERT method for finetuning performs well in cases where salient information is available beyond context while maintaining reasonable performance on complex classification tasks. Its performance indicates the potential benefits of using chunk-based approaches to capture important information beyond the context length.

\section{Limitation}

ChunkBERT assumes that a longer context benefits the downstream task, although this assumption may not hold universally. Some tasks may benefit from understanding longer contexts, while others may be less sensitive to longer inputs. In our experiments, we use a fixed chunk size of 128 tokens. However, the optimal chunk size may vary depending on the task and the datasets, finding the right balance in chunking granularity remains an important consideration and a potential area for future exploration within the ChunkBERT approach.
During training, ChunkBERT has a maximum input token limit of 4096 tokens. While this is sufficient for the datasets in this study, it imposes a fixed input length during training. Recent research\cite{DBLP:journals/corr/abs-2304-11062} suggests that learning and extrapolating to longer inputs during test time can be achieved with up to seven chunks. The ChunkBERT methodology can be used with a different chunk size during test time, for example, training with a chunk size of 128 tokens and performing inference with a full chunk size of 512 tokens. However, our preliminary experiments suggest that increasing the chunk size during test time may degrade performance. Further work is required to stabilize the extrapolation of chunk size during test time.

\section{Conclusion}
In this work, we propose a chunking finetuning approach called ChunkBERT for pretrained BERT models, which allows them to process inputs beyond 512 tokens to any arbitrary length. The method is flexible and can be easily applied from a standard GPU to any pretrained BERT model. It divides long input into shorter chunks and induces chunk representation using BERT. It further processes them using TextCNN resulting in linear memory growth with respect to chunk size. The proposed method is evaluated on a long text benchmark with various long document classification tasks. ChunkBERT performs 5.7\% better than LongFormer on average across the dataset and 18\% better on complicated datasets. The previous baselines display strong performance, but ChunkBERT still outperforms them marginally when evaluated only on the long samples from the test set while using only a fraction (6.25\%) of the original memory. Our findings suggest that ChunkBERT strikes a balance when a long-sequence modeling strategy depends on the dataset's complexity, information density, and the location of the salient information required for the task.

\section{Future Work}
There are multiple future avenues for this work, including studying the effect of changing chunk size on the performance of downstream tasks. In this work,  we only explore certain window sizes for the TextCNN module. Future work may involve comprehensive hyperparameter tuning of the TextCNN module and chunking parameters like the maximum number of allowed chunks. Here, we only explore TextCNN as an aggregation layer for routing inter-chunk information. We can explore an attention-based aggregation layer or grouped permutation of inputs for mixing chunk information. The choice of modeling strategy for long texts should consider the dataset's characteristics, the location and density of discriminative information, and the trade-offs between dense attention and context length. The experimental findings provide valuable insights into the performance variations and highlight the need for further research on benchmark datasets to enhance our understanding of long-text modeling.

\section*{Acknowledgement}
We acknowledge the support of the Natural Sciences and Engineering Research Council of Canada (\href{https://www.nserc-crsng.gc.ca/}{NSERC}) and the Canadian Institutes of Health Research (\href{https://cihr-irsc.gc.ca/}{CIHR}).
This research was enabled in part by support provided by ACENET (\href{https://ace-net.ca/}{ace-net.ca}) and the Digital Research Alliance of Canada (\href{https://alliancecan.ca/}{alliancecan.ca}). We thank the reviewers for their helpful comments on the draft. We thank Chandramouli Shama Sastry for implementing vectorized chunking and exciting research discussions about future directions. 

\bibliography{neurips_2023}
\bibliographystyle{plain}

\newpage
\appendix
\label{sec:appendix}
\section{Results on complete test set}
\label{appendix:a}
Table \ref{completetest} shows the results evaluated on the complete test set. These results show similar observations to the results on the long samples of the test set. The relative performance to the vanilla BERT model is shown in Figure \ref{completetest}.
\begin{table}[!htb]
    \centering
    \caption{Evaluation metrics on the complete test set for all datasets. The average accuracy(\%) over five runs is reported for Hyperpartisan and 20NewsGroup while the average micro-F1 is reported for other datasets. The highest value per column is in bold. \textit{Note}: The rows apart from ChunkBERT are reported from \cite{DBLP:journals/corr/abs-2203-11258} }
    \label{completetest}
    \resizebox{\textwidth}{!}{%
    \begin{tabularx}{\textwidth}{@{}lXXXXXX@{}}
    \toprule
           Model & Hyper-partisan & 20News-Group & EURLEX & Inverted-EURLEX & Book-Summary & Paired-Summary  \\
        \midrule 
        BERT & 92.00 & 84.79 & 73.09 & 70.53 & 58.18 & 52.24 \\ 
        BERT+TextRank & 91.15 & 84.99 & 72.87 & 71.30 & 58.94 & 55.99 \\ 
        BERT+Random & 89.23 & 84.65 & \textbf{73.22} & \textbf{71.47} & \textbf{59.36} & 56.58 \\ \midrule
        LongFormer & \textbf{95.69} & 83.39 & 54.53 & 56.47 & 56.53 & \textbf{57.76} \\ 
        ToBERT & 89.54 & \textbf{85.52} & 67.57 & 67.31 & 58.16 & 57.08 \\ 
        CogLTX & 94.77 & 84.63 & 70.13 & 70.80 & 58.27 & 55.91 \\ \midrule
        \textbf{ChunkBERT} & \textbf{95.69} & 84.46 & 70.92 & 69.49 & 54.08 & 56.42 \\ 
        \bottomrule
    \end{tabularx}%
}   \vspace{1pt}
\end{table}
\begin{figure}[!h]
    \centering
    \includegraphics[width=0.6\textwidth]{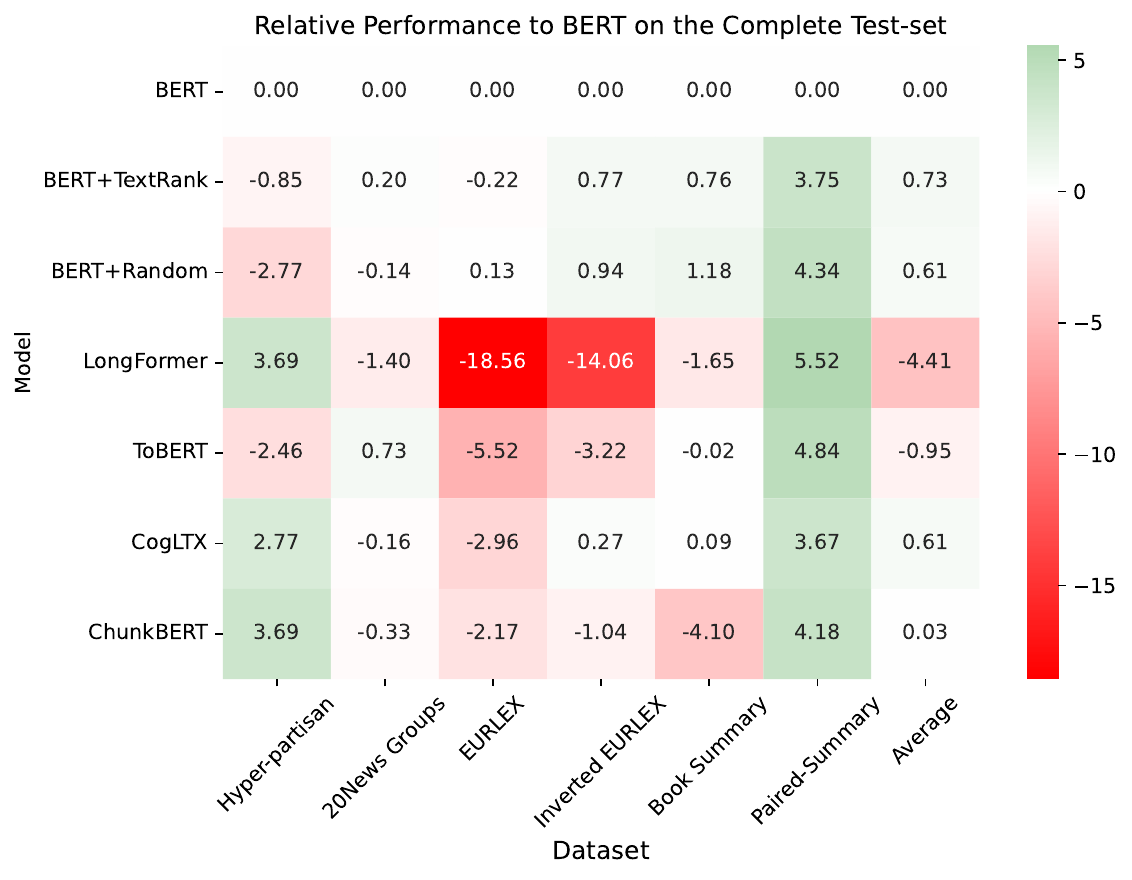}
    \caption{A heatmap displaying the performance metric difference ($\Delta$) between the BERT model and other models over the complete test set. It provides a visual comparison of performance gain or loss relative to BERT.}
    \label{fig:relative-performance-complete}
\end{figure}

\newpage
\section{Pseudo-code for vectorized chunking}
\label{appendix:b}
\begin{python}
from functools import partial 
import torch
def get_embedding(self,input_ids, attention_mask, token_type_ids):
    
    batch_size, n_tokens = input_ids.shape
    n_chunks = int(max(n_tokens/self.chunk_size,1))      
    # n_chunks * Batch * tokens 
    argument_chunks = [torch.cat(split_into_chunks(argument),dim=0) for argument in (input_ids,attention_mask,token_type_ids)]                   
    # condition: token_length//chunk_size == 0
    argument_chunks =[argument.reshape(n_chunks*batch_size,min(n_tokens,self.chunk_size)) for argument in argument_chunks] 
    
    # (n_chunks * Batch) * tokens * embed_size
    chunk_hidden_states = bert(argument_chunks[0],argument_chunks[1],argument_chunks[2],output_hidden_states=True)[2] 
                                    
    chunk_embed2d = torch.stack(chunk_hidden_states)[-5:].mean(0)      
    
    chunk_embed2d = torch.cat(chunk_embed2d.chunk(n_chunks),dim=1)
    
    return chunk_embed2d

def split_into_chunks(self):
    return partial(torch.split,split_size_or_sections=chunk_size,dim=1)

\end{python}

\end{document}